\documentclass{svproc}
\usepackage{url}

\usepackage{graphics} 
\usepackage[pdftex]{graphicx} 

\usepackage{amsmath} 

\usepackage{subfigure} 
\usepackage{paralist}
\usepackage{cite}

\usepackage{enumerate}
\usepackage{listings}

\usepackage[font=footnotesize]{caption}

\usepackage{enumitem}

\begin{document}
\mainmatter              
\title{Speed Maps: An Application to Guide Robots in Human Environments}
\titlerunning{Guide Robot}  
%
\author{Akansel Cosgun}
\authorrunning{Cosgun et al.} 
%
\tocauthor{Ivar Ekeland, Roger Temam, Jeffrey Dean, David Grove,
Craig Chambers, Kim B. Bruce, and Elisa Bertino}
\institute{Monash University, Australia\\
\email{akansel.cosgun@monash.edu}}

\maketitle              

\begin{abstract}

We present the concept of speed maps: speed limits for mobile robots in human environments. Static speed maps allow for faster navigation on corridors while limiting the speed around corners and in rooms. Dynamic speed maps put limits on speed around humans. We demonstrate the concept for a mobile robot that guides people to annotated landmarks on the map. The robot keeps a metric map for navigation and a semantic map to hold planar surfaces for tasking. The system supports automatic initialization upon the detection of a specially designed QR code. We show that speed maps not only can reduce the impact of a potential collision but can also reduce navigation time.

\end{abstract}


\section{Introduction}

Efficient collision-free navigation has been the longstanding goal of mobile robot navigation. As mobile robots are used more and more in human environments, they would need to exhibit interactive behaviors as well, such as following a person and guiding a person to a location. Such robots will be useful in a various type of settings. For example, in an airport the robot can carry the luggage of a passanger in amusement parks the robot can guide people to a attraction venues. Robots should demonstrate safe and socially acceptable navigation behaviors near humans. Recent research offer different approaches on this problem, however many of those approaches focus on finding only the path. In addition to the path, speed of mobile robots should also be dependent on the context. For example, a robot should move slowly and carefully in a hospital room while it is acceptable to navigate with high speed in a an empty corridor.


In this work, we present a one-time setup robotic system that exhibits social awareness for navigation. The setup requires an expert to map the environment with a robot and place speed limits according to the context. Any robot that detects a specially designed QR code can acquire this knowledge about the environment. We describe a system that can track people, follow them, and guide them to locations.  These locations are specified and labeled by the user. The robot calculates a goal point towards the landmark or waypoint and navigate under the speed limits. The contributions of this paper are 1) a method to guiding a person to a location and 2) the introduction of ``speed maps" for standardization of allowable robot speeds in human environments.

\section{Related Works}

In this section, we briefly review existing literature on localization, person detection/tracking, and human-aware navigation. 

\vspace{-0.5cm}
\subsection{Mapping and Localization}
\label{sec:rel1}

In mobile robotics, the standard practice for mapping and localization is described as follows: When the robot is first taken to a new environment, it has to map the environment. There has been extensive research on Simultaneous Mapping and Localization (SLAM) literature. The usual output is a binary 2D grid map where 1's represent obstacles and a 0's represent free space. Once the map is created, the robot can localize itself in the map using on-board sensing. Every time the robot is restarted, it has to start with an initial estimation of its location. Although there are global localization methods developed in the community, the usual practice is that the robotics expert manually provides an approximate initial location of the robot, then the localization method iteratively corrects the localization estimation as the robot moves in the environment.

\vspace{-0.5cm}
\subsection{Person Detection and Tracking}
\label{sec:rel2}

Using laser scanners for detecting and tracking people is common practice in robotics. Early works by \cite{montemerlo2002conditional} and \cite{schulz2001tracking} focused on tracking multiple legs using particle filters. Legs are typically distinguished in laser scans using geometric features such as arcs \cite{xavier2005fast} and boosting can be used to train a classifier on a multitude of features \cite{arras2007using}. Topp \cite{topp2005tracking} demonstrates that leg tracking in cluttered environments is prone to false positives. For robust tracking, some efforts fused information from multiple lasers. Carballo\cite{Carballo2008} uses a second laser scanner at torso level. Glas \cite{glas2009laser} uses a network of laser sensors at torso height in hall-type environments to track the position and body orientation of multiple people. Several works used different modalities of sensors to further improve the robustness. Some applications \cite{kleinehagenbrock2002person,bellotto2009multisensor}  combine leg detection and face tracking in a multi-modal tracking framework.

\vspace{-0.5cm}
\subsection{Navigation in Human Environments}
\label{sec:rel3}

Among the early works of guide robots, museum tour-guide robots are the most common \cite{nourbakhsh2003mobot, thrun1999minerva}. Burgard~\cite{burgard1999experiences} talks about the experiences obtained in a museum obtained over six days. Bennewitz \cite{bennewitz2005towards} focuses on interaction with multiple people for museum robots. Pacchierotti \cite{pacchierotti2006design} presents deployment of a guide robot in an office environment and analyses interactions with bystanders. Kanda \cite{kanda2009affective} desribes a guide robot that was deployed in a shopping mall for 25 days. Some works considered evacuation scenarios where robots need to guide people to safety \cite{kim2009portable , robinette2011incorporating}.

Person following has been extensively studied the literature. Loper \cite{loper2009mobile} presents a system that is capable of responding to verbal and non-verbal gestures and follow a person. A common method to follow a group of people is to choose and follow a leader \cite{stein2013navigating}. Some papers addressed the problems encountered during following, such as robot speed not catching up with the person \cite{cai2013robot} and how to recover when the robot loses track of the person \cite{ota2013recovery}.


\section{Elements for Navigation}
\label{sec:navigation}

In this section, we provide an overview of our navigation system. During regular operation, the robot waits at a base location for new users. When a robot gets close to a robot, the robot rotates towards the user and listens for input from the GUI. User can either send the robot to a previously saved waypoint or planar landmark, or have the robot guide him/her to that location. We use Robot Operating System (ROS) Navigation stack for basic point-to-point navigation and provide different goal points according to the task. The speed of the robot is adjusted using the context of the environment. For example, near blind corners and in small rooms the robot's speed is limited, however in open corridors robot is allowed to navigate faster. Our system requires only a one time setup by a robotics expert. A navigation-capable robot that is restarted or coming to the environment for the first time can acquire map of the environment whenever it reads a special QR-code that is placed to a known location in the environment. The QR code includes links to the map and landmarks, speed limits, as well as the location of the QR pattern in the map. Therefore the robot automatically can infer its initial location upon detection of the QR code, without external intervention. The environment is initially mapped during operation using \textit{gmapping} SLAM package and robot is localized by the \textit{amcl} package of ROS.

This section is organized as follows. We present the state machine used for this work in Sec.~\ref{sec:state_machine}, followed by the brief description of the robot maps in Sec.~\ref{sec:maps} and tablet GUI in Sec.~\ref{sec:gui}. The robot needs global initialization the first time it is started. In Sec.~\ref{sec:qr}, we present our QR-base approach and how the robot finds its own position upon detection of the QR code. Then we discuss how the goal locations for guidance is determined in Sec. \ref{sec:goal_locations}. When a plane is labeled instead of a waypoint, the robot first needs to find a goal location that is nearby the labeled landmark. In Sec.~\ref{sec:speed_maps}, we introduce the speed maps, which gives the robot awareness about how fast it is permitted to navigate. The speed maps are calculated off-line, and is dependent on the context.

\begin{figure}[ht!]
\centering
\includegraphics[width=0.9\textwidth]{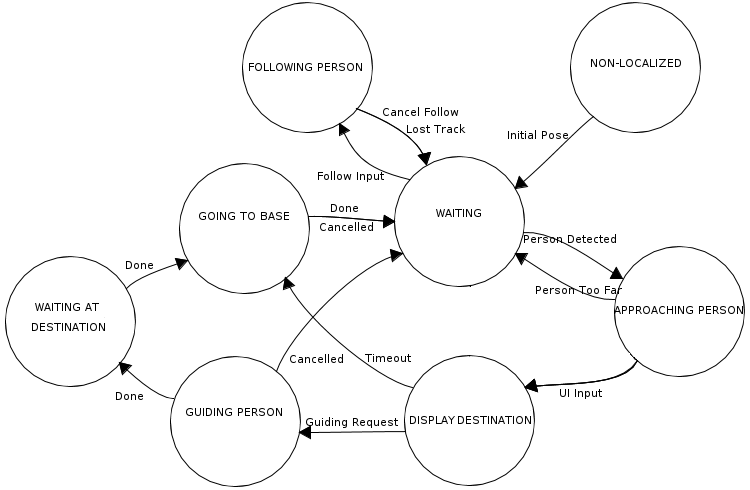}
\caption{Finite State Machine for the Guide Robot}
\label{fig:fsm}
\end{figure}

\subsection{State Machine}
\label{sec:state_machine}

The robot's higher level actions are governed by a finite state machine (Figure~\ref{fig:fsm}). When the robot is first initialized, it is in the \textit{NON-LOCALIZED} state. When the initial pose is given, either manually or by detection of the QR code, robot switches to \textit{WAITING} state. This is the state where the robot actively looks input from users. When a person is closeby, robot approaches towards the person so its tablet GUI faces the person. If the person inputs a guide request, the robot displays destination and asks for confirmation. If confirmed, the robot switches to the \textit{GUIDING PERSON} state. When guiding is completed, robot waits for a while and goes back to the base. At any time, a user can cancel the operation and enter new commands.

\subsection{Maps}
\label{sec:maps}

We use both a metric and a semantic map. To generate semantic maps, we use our previous work \cite{trevor2012maplabeling}. Maps that include semantic information such as labels for structures, objects, or landmarks can provide context for navigation and planning of tasks, and facilitate interaction with humans. Planar surfaces can be detected from point cloud data, and represented in a consistent map coordinate frame. Maps composed of such features can represent the locations and extents of landmarks such as walls, tables, shelves, and counters. We represent planes by its convex hull's 3D coordinates. 

\subsection{User Interface}
\label{sec:gui}

\begin{figure}[ht!]
\centering
\includegraphics[width=0.75\textwidth]{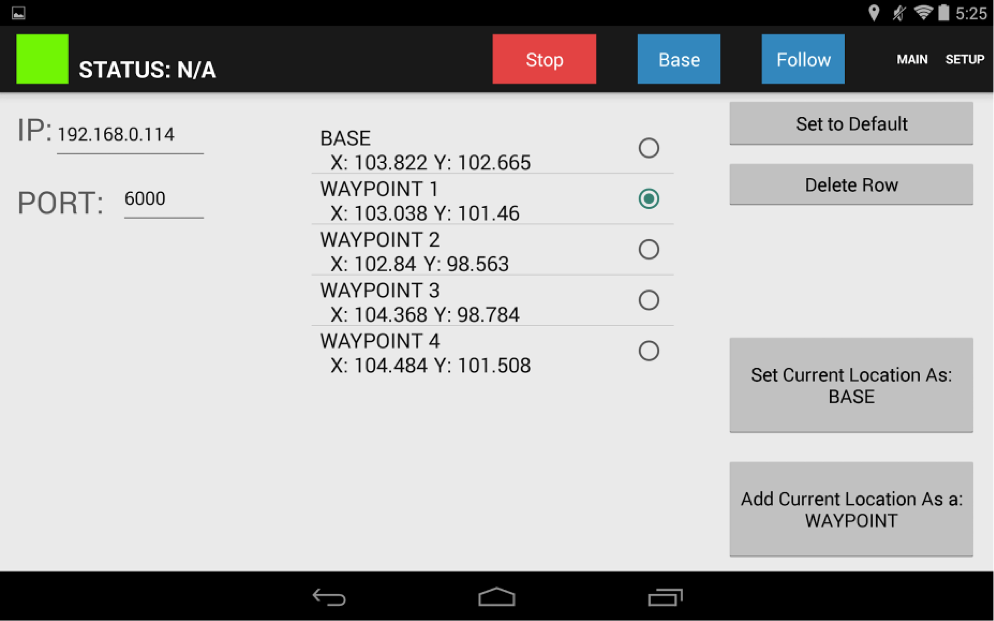}
\caption{Tablet GUI for commanding the robot to navigate to various goal positions}
\label{fig:ui}
\end{figure}


Using an on-board tablet interface, the user can also enable person following. A screenshot from the GUI for saving waypoints is shown in Figure~\ref{fig:ui}.

\subsection{Global localization}
\label{sec:qr}

This section describes how robots can acquire knowledge of the environment without doing the mapping themselves. The proposed system can be used for 2 purposes:

\begin{enumerate}
\item A robot enters to an environment the first time and does not have the map
\item Robot has the map but does not know its initial position in the map
\end{enumerate}

System designers can strategically place QR code tags to the walls/floors in the environment. These places can be walls at the entrance of an indoor space, or mostly visited areas. QR codes in general contain links, text and various other data. In our application, the QR pattern includes data in XML format, which includes the link to the map and speed map files, and the position of the QR pattern in the map. 

\begin{figure}[ht!]
\centering
\includegraphics[width=0.6\textwidth]{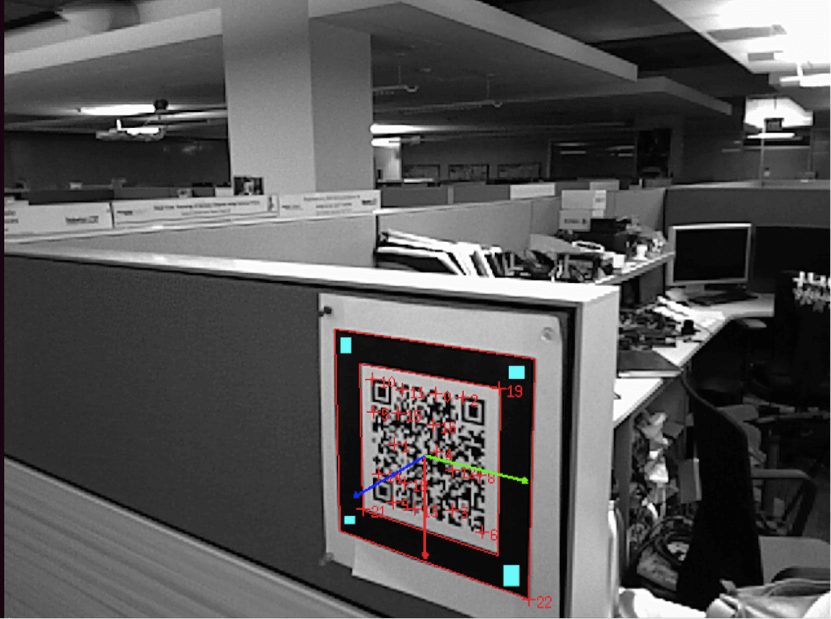}
\caption{A robot can acquire map information and localize itself against the map upon detection of a specially designed QR code}
\label{fig:qr}
\end{figure}


The position provided in the pattern is the position and orientation (in quaternions) of the QR tag in the map frame. Using this information, a robot locates itself in the map, where:

$^{robot}T_{cam}$ is the transformation from the robot base frame to the camera frame on the robot and is known.

$^{cam}T_{QR}$ is the pose of the QR code in the camera frame and is available upon detection of the QR code.

$^{map}T_{QR}$ is the transformation from the robot base frame to the camera frame on the robot and is read from the data embedded to the QR code.

$^{map}T_{robot}$ is the pose of the robot in the map and is unknown.\\

$^{map}T_{robot}=(^{map}T_{QR})*(^{QR}T_{robot})$

$^{QR}T_{robot}=(^{QR}T_{cam})*(^{cam}T_{robot})=(^{cam}T_{QR})^{-1} * (^{robot}T_{cam})^{-1}$

Therefore:

$^{map}T_{robot}=(^{map}T_{QR})*(^{cam}T_{QR})^{-1} * (^{robot}T_{cam})^{-1}$\\

$(x,y,\theta)$ of the robot in the map frame is required for initialization. Position $(x,y)$ is readily found by looking at the displacement of the transformation. Orientation $\theta$ is found by projecting the axes to the map floor plane and . After the initial pose is provided, \textit{amcl} package handles the localization using the laser scanner readings.

We use the \textit{visp\_auto\_tracker} ROS package to extract the pose of the QR code and the content embedded in the pattern. To interpret the XML data, we use libXML++ package.

\subsection{Goal Location Determination}
\label{sec:goal_locations}

We want the robot to be able to navigate to semantic landmarks. If the label is attached to an explicit coordinate (using the current position of the robot during labeling), then goal is readily that coordinate.

We consider two cases when calculating a goal point for a given label. The requested label may correspond to one landmark, or multiple landmarks. In the case that it is one landmark, it corresponds to a plane somewhere in the environment, but if multiple landmarks share the label, we assume the entity corresponds to a volume. For example, a user may label all four walls of a room, so the extent of the room is represented. The robot should infer that a volume is encoded with the labels provided to it, and find a proper goal location accordingly.

\subsubsection{Only one plane has the requested label}

\begin{figure}[ht!]
\centering
        \subfigure[]{%
        	\label{fig:fig:hallway1}
            \includegraphics[width=0.4\textwidth]{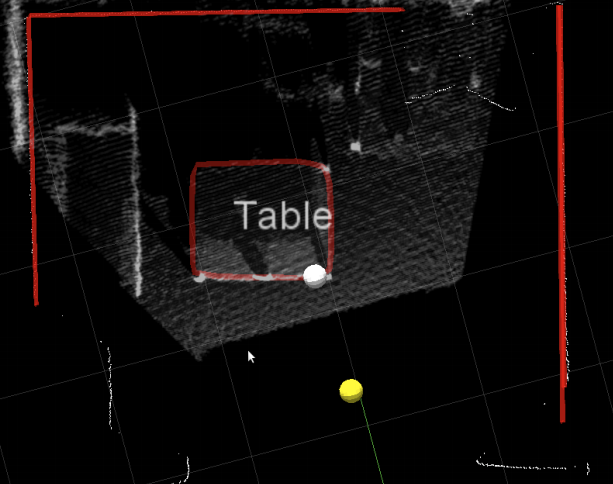}
        }
        \subfigure[]{%
                	\label{fig:hallway2}
           \includegraphics[width=0.47\textwidth]{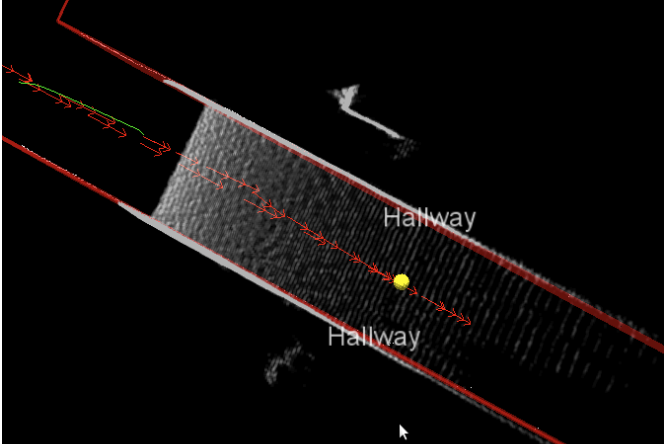}
        }        
    \caption{%
	Our system allows two types of semantic features: a) Planar surfaces and b) Rooms
    }%
   \label{fig:single_plane}
\end{figure}

In this case, we assume that the robot should navigate to the closest edge of the plane, so we select the closest vertex on the landmark's boundary to the robot's current position. We calculate a line between the closest vertex and the robot's current pose, and navigate to a point on this line 1 meter away from, and facing the vertex. This method is suitable for both horizontal planes such as tables, or vertical planes such as doors.

\subsubsection{Multiple planes has the requested label}

In this case, we project the points of all planes with this label to the ground plane, and compute the convex hull. For the purposes of navigating to this label, we simply navigate to the centroid of the hull. While navigating to a labeled region is a simple task, this labeled region could also be helpful in the context of more complex tasks, such as specifying a finite region of the map for an object search task.

\subsection{Speed Maps}
\label{sec:speed_maps}

\begin{figure}[ht!]
\centering
\includegraphics[width=0.9\textwidth]{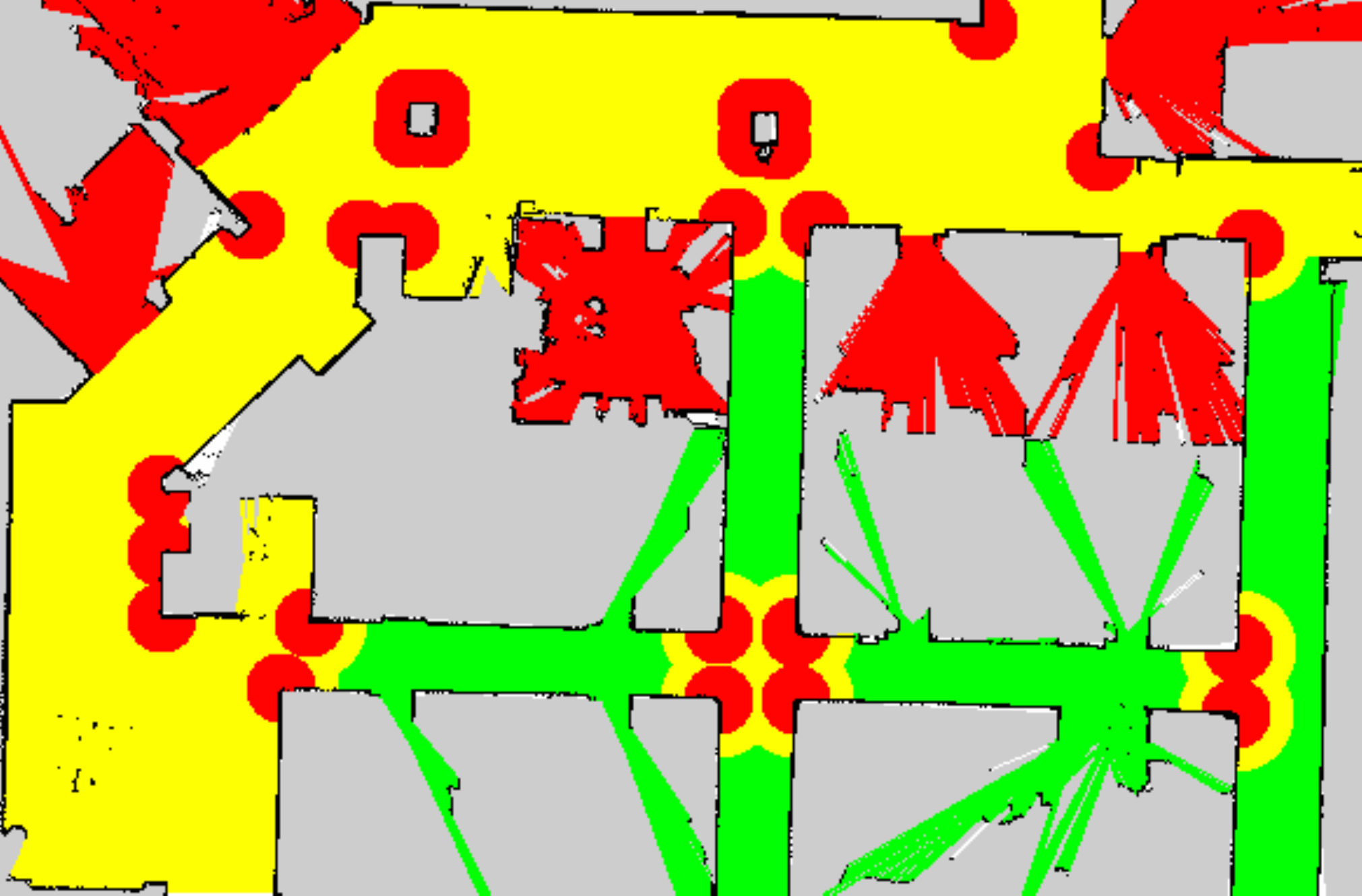}
\caption{Static speed map for an office environment. The robot has to be slow in red zones, can have moderate speed in yellow zones and can speed up to a limit in green zones.}
\label{fig:speed_map}
\end{figure}

In this section, we introduce speed maps, which imposes speed limits to robots. A speed map can be static, so that it defines speed zones as in Figure~\ref{fig:speed_map}, or it can be dynamic, so that the robot can adjust its speed around humans. We claim that usage of such speed maps can make the robots safer, more predictable and possibly more efficient. 

Speed maps can be used in a layered manner. At the bottom, there is the static speed map, which originated from the layout. Dynamic speed maps, which change with time, can build on top of the static speed map, for example impose speed constraints around humans.

We define three zones for static speed maps: 

\begin{itemize}
\item Green zone: It is safe to speed up
\item Yellow zone: Relatively safe, but human interaction possible
\item Red zone: Human encounter likely, move slowly
\end{itemize}

The speed map shown in Figure~\ref{fig:speed_map} depicts an office environment. This speed map is designed by hand using the following rules: Spaces corresponding to rooms and cubicles are covered as Red Zones. Blind corners are covered with a Red Zone close to corner and Yellow zone enclosing the Red Zone. Corridors are covered as Green Zones and the rest is covered as Yellow Zones. Although we manually labeled this speed map, there are several approaches in the literature that does automatic room categorization. Room segmentation has been proposed in an interactive fashion by Diosi \cite{diosi2005interactive}, as well as automatically, especially for creating topological maps \cite{mozos2007supervised}.

An example for a dynamic speed map is given in Section ~\ref{sec:person_guidance} and will be discussed context of speed control for guidance.

\section{Implementation}
\label{sec:interactive_navigation}

In this section, we discuss how the following and guidance behaviors are obtained for a mobile robot. In Section~\ref{sec:person_tracking}, we describe how the multimodal person detection and tracking is performed. In Section~\ref{sec:person_following} we discuss the person following behavior and in Section~\ref{sec:person_guidance} we discuss person guidance.

\subsection{Person Detection and Tracking}
\label{sec:person_tracking}

In order to interact with people, the robot has to be able able to track them. We use a person tracking system that fuses detections from two sources. These detectors are the leg detector and the torso detector. The reason for using multiple detectors for the person tracking system is to to increase the robustness of the tracking and provide 360 degrees coverage around the robot. Below are the brief descriptions of the person detection methods:

\subsubsection{Leg Detector}

Ankle-height front-facing laser scanner (Hokuyo UTM 30-LX) is used for this detector. We trained a classifier for detection using three geometric features: Width, Circularity and Inscribed Angle Variance~\cite{xavier2005fast}. We find a distance score for each segment in the laser data using the weighted sum of the distance to each feature and threshold the distance for detection.

\subsubsection{Torso Detector}

Torso-height back-facing laser scanner (Hokuyo UTM 30-LX) is used for this detector. We model the human torso as an ellipse and fit each segment in the laser image an ellipse by solving the problem with a generalized eigensystem~\cite{fitzgibbon1999direct}. Then the axis lengths of the fitted ellipse is compared to ellipse parameters obtained from the training data. More information on this detector is provided in~\cite{cosgun2014guidance}.

\subsubsection{Person Tracking}

The detections from these different sources are fused using a Kalman Filter with constant velocity model. We used a simple Nearest Neighbors approach for associating tracks with detections. Whenever a track is not observed for a couple of seconds, it is deleted from the list.

\subsection{Person Following}
\label{sec:person_following}

When the robot is following a person, it constantly aims to keep a fixed distance from the person. In our experiments, we found that keeping 0.9 meters was far enough for the person to feel safe, but close enough to keep track of the target. The robot updates its goal position that is following distance away from the target at every control step (10Hz). This goal point is fed to the ROS Navigation Stack, which calculates the velocity commands to keep the robot on the path. If the goal position happens to intersect an obstacle, a new goal position is found by raytracing towards the robot and finding the first non-colliding position. 

\subsection{Person Guidance}
\label{sec:person_guidance}

In previous work, we developed a system to guide blind individuals to a goal position using a haptic interface \cite{cosgun2014guidance} . In this work we describe a more generalized approach. To guide a person to a location, the robot has to plan a path for itself and try to keep the person engaged in the guidance task. This actually is a hard problem if the robot also plans a separate path for the user, however we use a simpler approach for guidance, which produces plausible results. The robot executes the path that is calculated by ROS navigation, however modulates its speed according to the distance to the user. This actually corresponds to a dynamic speed map, that is used in conjunction with the static speed map described in Section~\ref{sec:speed_maps}.

\begin{figure}[ht!]
\centering
\includegraphics[width=0.4\textwidth]{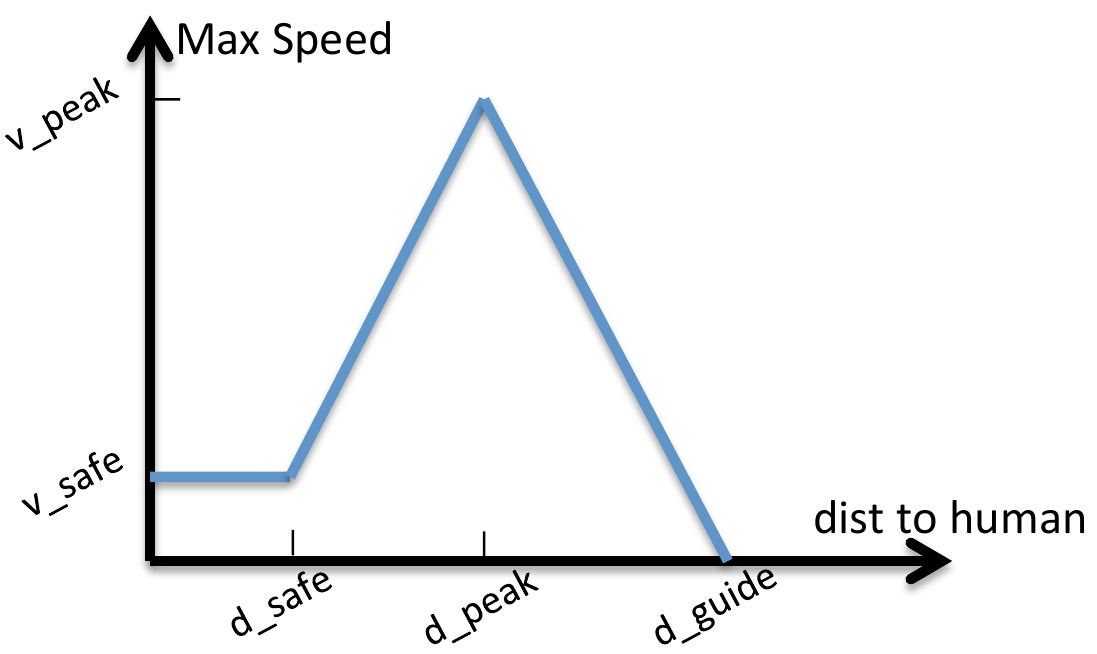}
\caption{Speed profile of the robot for guiding a person.}
\label{fig:profile}
\end{figure}

We define a speed profile function that is a function of the distance between the robot and the human (Figure \ref{fig:profile}). The robot moves at a low speed $v_{safe}$ if the human is too close. The speed is peaked at distance $d_{peak}$ and the robot stops if the distance is larger than $d_{guide}$, which may indicate that the human is not interested in being guided. Note that this speed profile is subject to the static speed limits, i.e. $v_{peak}$ is capped by the static speed map limit.

\section{Demonstration}
\label{sec:results}

\begin{figure}[ht!]
\centering
\includegraphics[width=0.2\textwidth]{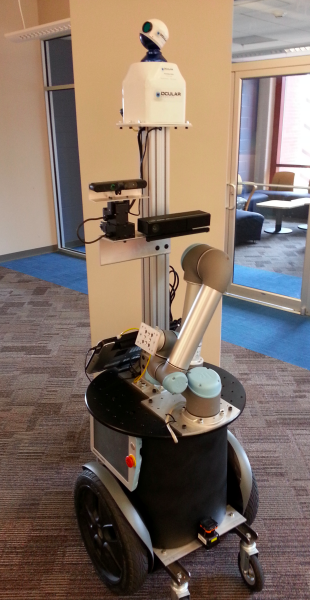}
\caption{The Segway robot platform used in our experiments}
\label{fig:jeeves}
\end{figure}

The system is implemented on a Segway RMP-200 robot with casters (Figure~\ref{fig:jeeves}). We made use of ROS Architecture, for mapping, localization and path planning. We evaluate the system in two scenarios: corridor navigation and guidance.

In the corridor navigation scenario, the robot has a goal point and it is not guiding anybody. In the first condition, we used standard ROS Navigation and in the second condition, we used ROS Navigation with our speed modulation method described in the static speed maps section. There is a person standing right around the corner, which is invisible to the robot until it actually starts turning around the corner. Evaluation criteria is the execution time and the actual robot positions along the path.

In the guidance scenario, there is no person around the corner and the robot guides the person to the same goal. In the first condition, we use ROS Navigation but robot stops if the distance to the human is over a threshold. In second condition, dynamic speed method is used. Instantaneous speed of the robot and the human are logged for comparison.

\subsection{Scenario: Point-to-point navigation}

\begin{figure}[ht!]
\centering
        \subfigure[]{%
        	\label{fig:nav_black}
            \includegraphics[width=0.47\textwidth]{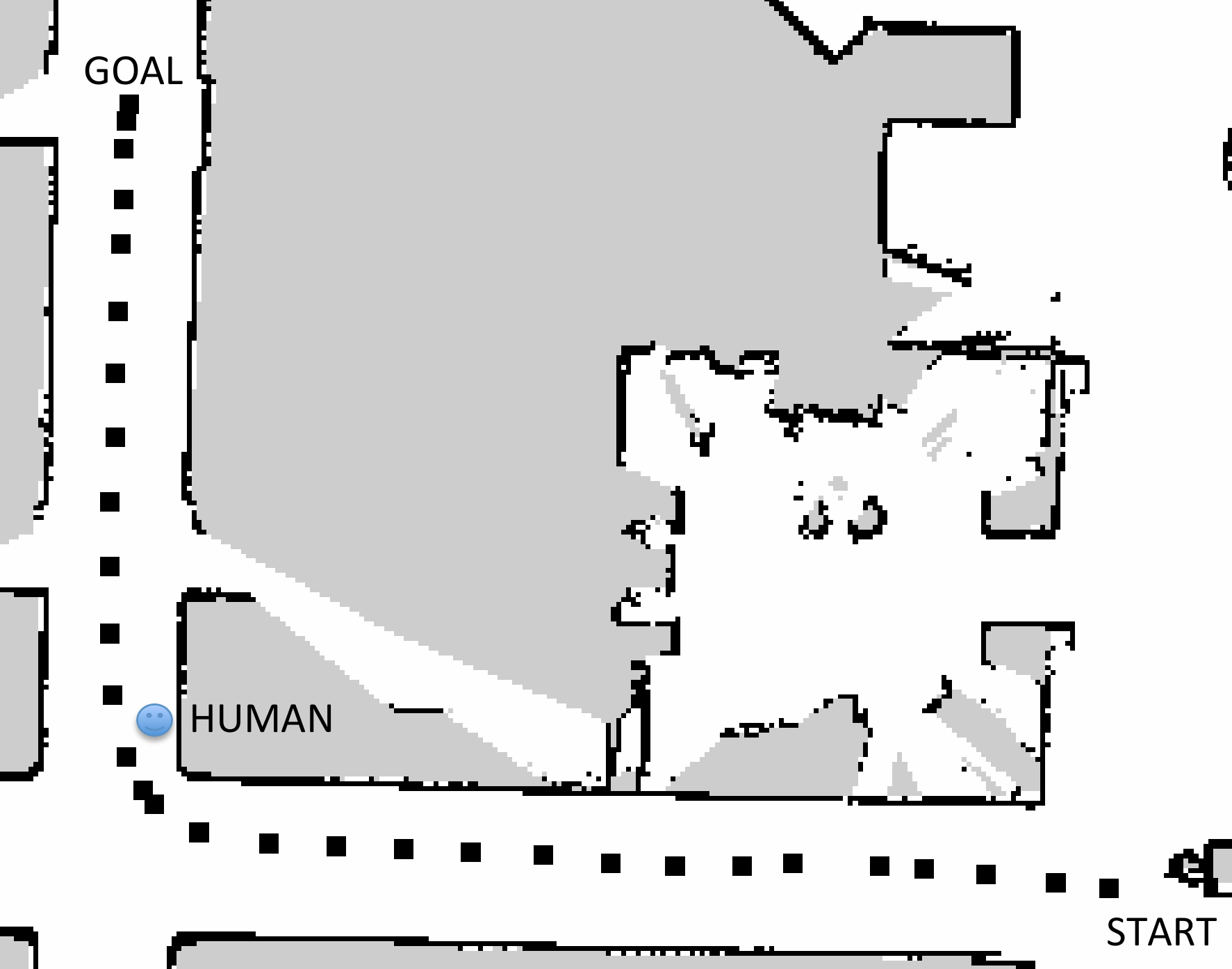}
        }
        \subfigure[]{%
                	\label{fig:nav_color}
           \includegraphics[width=0.47\textwidth]{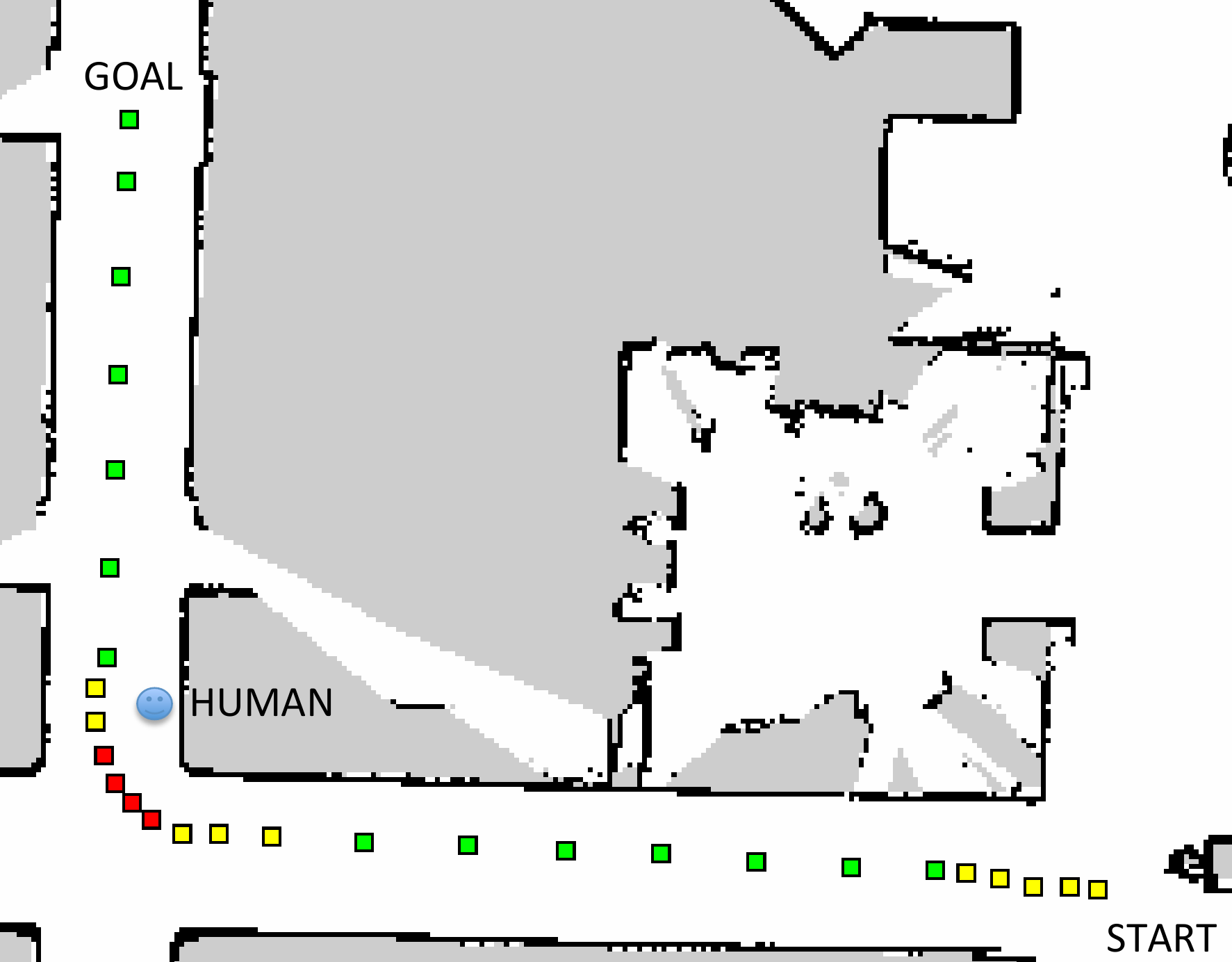}
        } 
    \caption{%
    Corridor scenario. Robot is given a fixed goal location, and it is not guiding a user. There is a bystander person standing right around the corner, that is not visible to the robot until it turns the corner. Points annotate robot position taken at fixed time intervals. 
	a) ROS Navigation. Note that the distance between robot positions are mostly constant. The robot gets very close to the bystander because it is moving relatively fast when it turned the corner b) Our approach using static speed map. Green points annotate Green Zone where robot can move fast, yellow points are where robot is limited to moderate speed, and red points are when the robot is in Red Zones and has to move relatively slow. It can be seen by looking at robot's positions that this approach was more gracious turning the corner and respecting human's personal space. It also reached the goal faster.
     }%
   \label{fig:nav_black}
\end{figure}

The speed map shown in Figure~\ref{fig:speed_map} is used for the experiments. The speed limits were set as: $v_{green}=1.5$, $v_{yellow}=0.5$, $v_{red}=0.15$, all m/s. We compared our approach with fixed maximum speed of 1.0 m/s. The paths taken by the robot are compared in Figure~\ref{fig:nav_black}. With our approach, the robot slowed down before turning, which allowed for early detection of the human and it was able to give more space to the human. Our approach was slightly more efficient, reaching the goal in 28s as opposed to 29.1s.

\subsection{Scenario: Guiding a person}

In this scenario, the robot is given a fixed goal to guide a person. We compared the velocity profile in Figure~\ref{fig:profile} with $d_{guide}=1.7$m, $d_{peak}=0.9$m, $d_{safe}=0.1$m, $v_{safe}=0.1$m/s, $v_{peak}=1.0$m/s. We compared our approach with the simple strategy: If the human is closer than $d_{guide}$, then the navigation continues with a fixed max speed. Otherwise robot stops and waits.

\begin{figure}[ht!]
\centering
        \subfigure[]{%
        	\label{fig:graph_noadj}
            \includegraphics[width=0.47\textwidth]{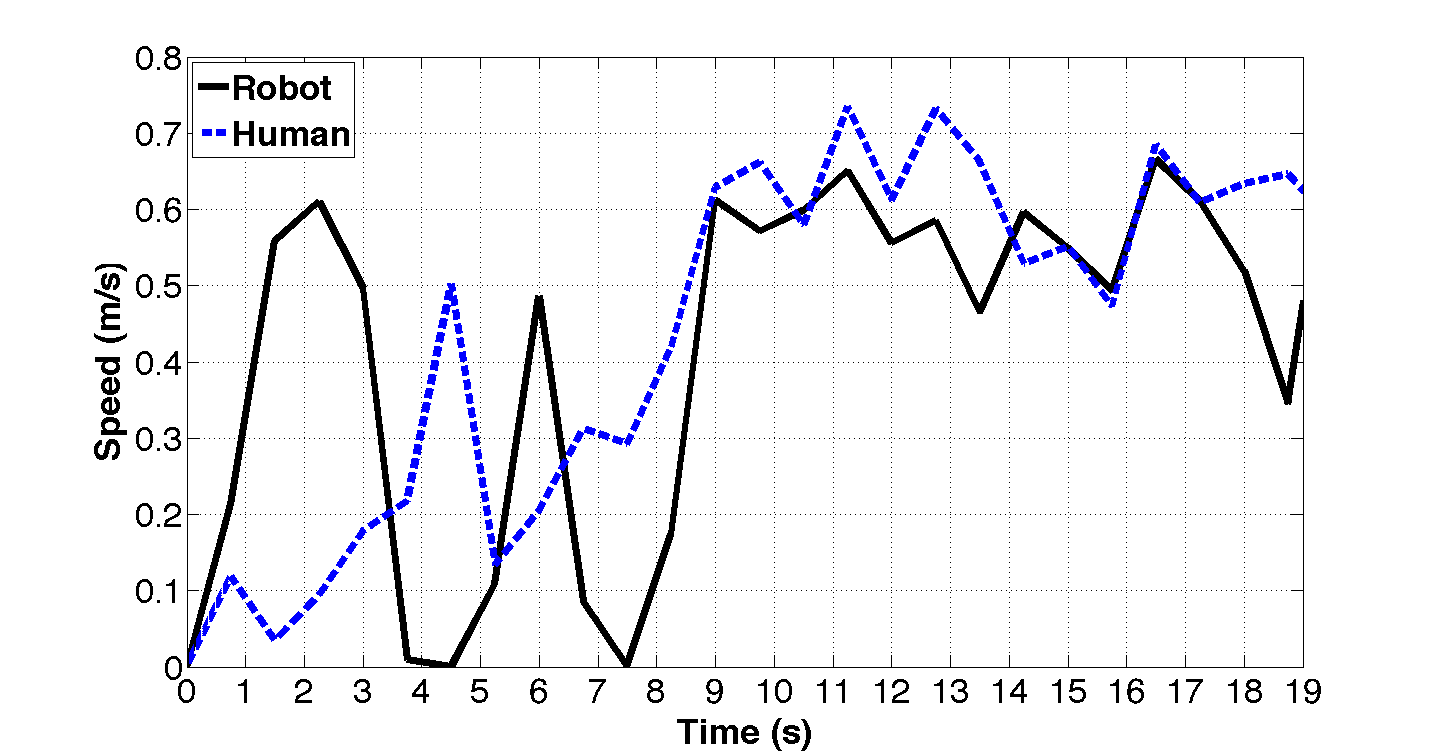}
        }
        \subfigure[]{%
                	\label{fig:graph_adj}
           \includegraphics[width=0.47\textwidth]{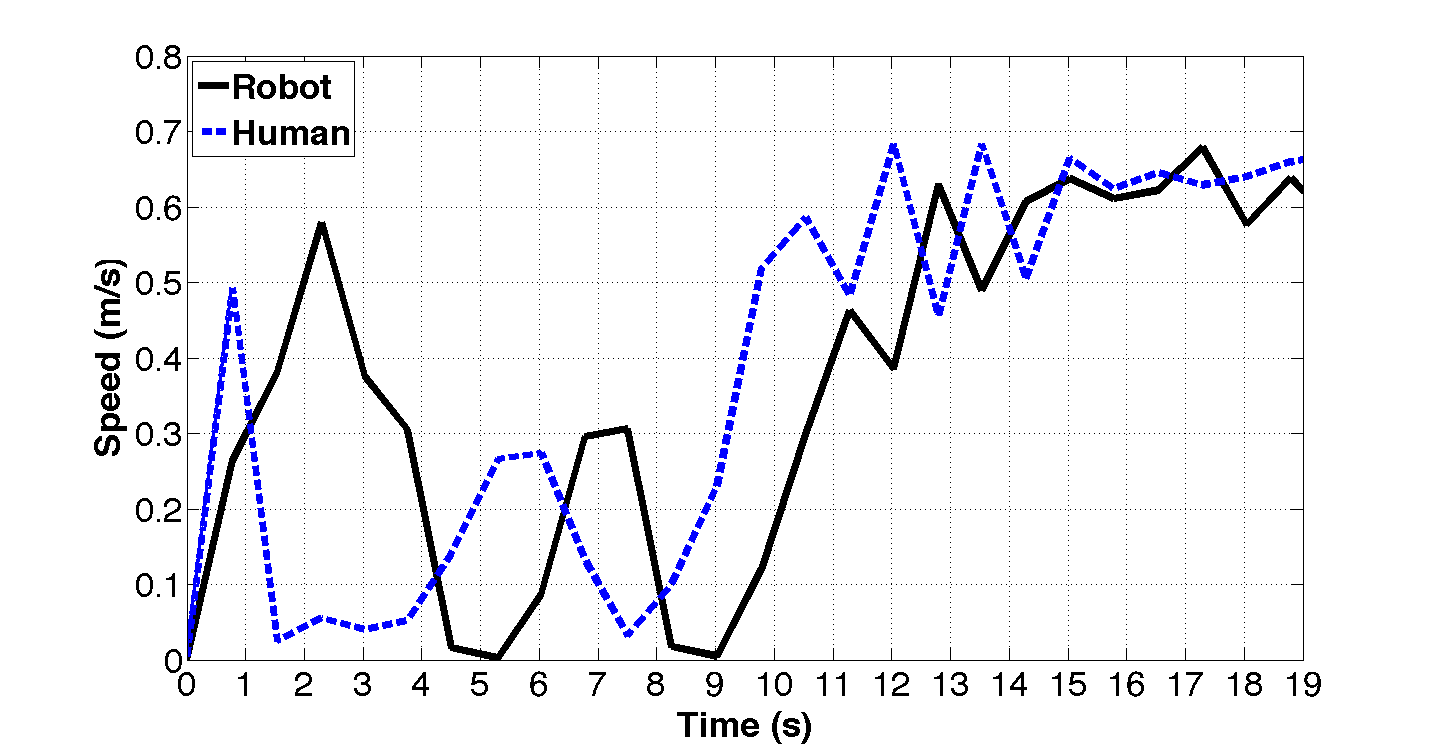}
        }        
    \caption{%
	Comparison of robot and human speeds with respect to time. a) Standard ROS Navigation b) Our approach: accelerations are less steeper than a), which employs the dynamic speed adjustment for guidance.
     }%
   \label{fig:graph}
\end{figure}

The comparison of robot speeds is given in Figures~\ref{fig:graph_noadj} for fixed max speed and~\ref{fig:graph_adj} with the speed profile method. Between $t=0$ and $t=9s$, the accelerations are steeper for the fixed max speed case. Robots that exhibit high accelerations will likely be perceived as unsafe, therefore our approach exhibits a more socially acceptable behavior. Moreover, after the person started following ($t >9s$), our approach is better at matching the speed pattern of the human.

\section{Conclusion}
\label{sec:conclusion}

In this work, we presented a context-aware robot that accepts user-annotated landmarks as the goal. By reading special QR codes, the robot acquires knowledge about the environment and its global position. We showed that speed maps not only can reduce the impact of a potential collision, but it can also reduce task execution time. Robots yielding to speed maps also can potentially be can perceived as more safe as the speed limits serves as traffic rules. We also showed that our guidance approach results in more gracious motions compared to standard ROS Navigation.

Future work includes the manual design of speed maps in Augmented Reality 
\cite{gu2021seeing}, incorporating the speed maps concept with human-aware navigation \cite{cosgun2016anticipatory}, perhaps for a different application such as a photographer robot \cite{newbury2020learning}, and conducting user studies to evaluate the usability of the system with metrics tailored for navigating among people \cite{wang2021metrics}.

\bibliographystyle{IEEEtran}

\bibliography{speed_maps}

\end{document}